\documentclass[lettersize,journal]{IEEEtran}
\usepackage{amsmath,amsfonts}
\usepackage{algorithmic}
\usepackage{algorithm}
\usepackage{array}
\usepackage[caption=false,font=normalsize,labelfont=sf,textfont=sf]{subfig}
\usepackage{textcomp}
\usepackage{stfloats}
\usepackage{url}
\usepackage{verbatim}
\usepackage{graphicx}
\usepackage{cite}
\usepackage{hyperref}
\usepackage{makecell}
\usepackage{pdflscape}
\usepackage{adjustbox}
\usepackage{enumitem}
\usepackage{multirow}
\usepackage{pifont}
\usepackage{tabularx}
\newcommand{\xmark}{\ding{55}}%
\newcommand{\cmark}{\ding{51}}%
\begin{document}
\title{TypeFormer: Transformers for\\Mobile Keystroke Biometrics}

% \author{ABC\thanks{XYZ} \and DEF\thanks{UVW} \and GHI\footnotemark[1]}

\author{\IEEEauthorblockN{Giuseppe Stragapede\IEEEauthorrefmark{1}\thanks{Email: \href{mailto:giuseppe.stragapede@uam.es}{giuseppe.stragapede@uam.es}}, 
Paula Delgado-Santos\IEEEauthorrefmark{2}\IEEEauthorrefmark{1}, 
Ruben Tolosana\IEEEauthorrefmark{1}, \\
Ruben Vera-Rodriguez\IEEEauthorrefmark{1},
Richard Guest\IEEEauthorrefmark{2} and
Aythami Morales\IEEEauthorrefmark{1}
}

\IEEEauthorblockA{\IEEEauthorrefmark{1}Biometrics and Data Pattern Analytics (BiDA) Lab, Universidad Autonoma de Madrid, Spain}

\IEEEauthorblockA{\IEEEauthorrefmark{2}School of Engineering, University of Kent, United Kingdom}
}


\maketitle

% \thanks{organization={School of Engineering},            addressline={University of Kent}, 
%             % city={},
%             % postcode={}, 
%             % state={},
%             country={United Kingdom}}



\begin{abstract}
The broad usage of mobile devices nowadays, the sensitiveness of the information contained in them, and the shortcomings of current mobile user authentication methods are calling for novel, secure, and unobtrusive solutions to verify the users' identity. In this article, we propose \textit{TypeFormer}, a novel Transformer architecture to model free-text keystroke dynamics performed on mobile devices for the purpose of user authentication. The proposed model consists in Temporal and Channel Modules enclosing two Long Short-Term Memory (LSTM) recurrent layers, Gaussian Range Encoding (GRE), a multi-head Self-Attention mechanism, and a Block-Recurrent Transformer layer. Experimenting on one of the largest public databases to date, the Aalto mobile keystroke database, TypeFormer outperforms current state-of-the-art systems achieving Equal Error Rate (EER) values of 3.25\% using only 5 enrolment sessions of 50 keystrokes each. In such way, we contribute to reducing the traditional performance gap of the challenging mobile free-text scenario with respect to its desktop and fixed-text counterparts.
Additionally, we analyse the behaviour of the model with different experimental configurations such as the length of the keystroke sequences and the amount of enrolment sessions, showing margin for improvement with more enrolment data. Finally, a cross-database evaluation is carried out, demonstrating the robustness of the features extracted by TypeFormer in comparison with existing approaches. 
\end{abstract}


\section{Introduction}
\label{sec:Introduction}
\begin{figure*}[t]
    \centering
     \includegraphics[trim={0cm 0cm 0 0cm}, width=0.9\linewidth]{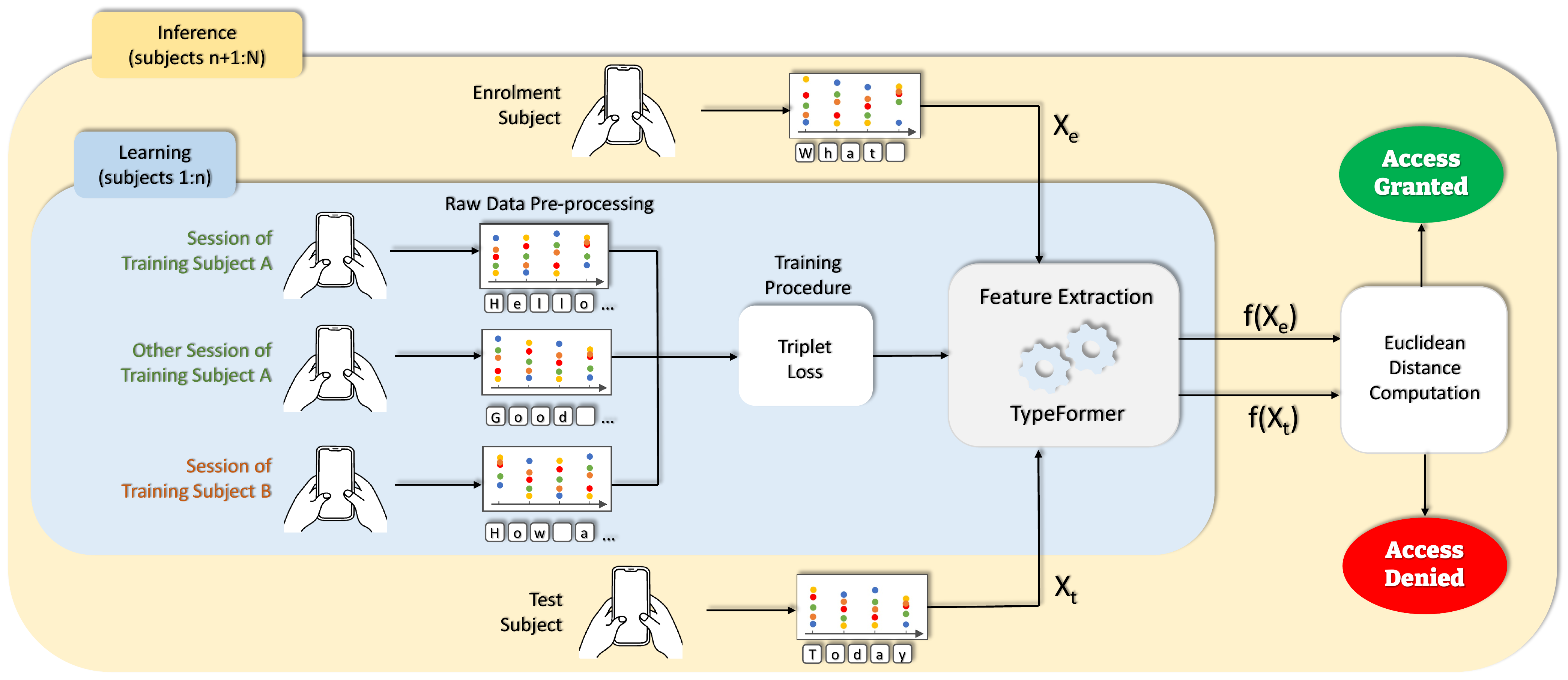}
    \caption{Graphical representation of the workflow of TypeFormer, the proposed biometric keystroke free-text verification system.}
    \label{fig:workflow}
\end{figure*}

\IEEEPARstart{T}{he} rapid digitalisation of the society, together with the pervasiveness of mobile devices, is making room for unprecedented Human-Computer Interaction (HCI) scenarios. Most people are now constantly connected to the internet through their mobile devices, accessing remotely their private data, and carrying out sensitive operations in sectors such as Banking, Financial Services and Insurance (BFSI), healthcare, e-commerce, and government, among many others \cite{THARIQAHMED2020103281}. This trend has increased the amount of cybercrimes observed \cite{rathgeb2022handbook}, evidencing the need for novel and reliable security methods that fulfill context-specific constraints, such as: \textit{(i)} continuous protection; \textit{(ii)} user-friendliness; \textit{(iii)} limited processing load, compatible with mobile environment specifications; \textit{(iv)} immunity to spoofing. To meet such requirements, recent studies have explored the feasibility of the user's behavioural\footnote{In contrast to \textit{physiological} biometrics, which pertains to the biological characteristics of an individual, such as face or fingerprint, all means that enable or contribute to differentiating between individuals throughout the way they perform activities are labelled as \textit{behavioural}, i.e., gait, keystroke dynamics, handwritten signature, etc.} biometric traits as an authentication method to create an additional transparent security layer on top of traditional approaches \cite{ISO2018, 7503170}. In fact, such traits can be constantly verified in a \textit{passive} way \cite{STRAGAPEDE202235, DELGADOSANTOS202230}, i.e., without having the user to carry out any specific \textit{entry-point} authentication task, such as placing their fingertip on the dedicated sensor, or typing a pass code, thus addressing \textit{(i)} and \textit{(ii)}. Such methods are also convenient as mobile devices come equipped with several sensors that can be treated as sources of biometric modalities \cite{delgadosantos2021survey, PORWIK2021104135}. Mobile behavioural biometric traits are also captured as low-dimensional time domain signals, i.e., the acquisition and processing is fast \textit{(iii)}. Additionally, it has been argued that spoofing behavioural biometrics requires more advanced technical skills compared to their physiological counterparts \textit{(iv)} \cite{rathgeb2022handbook}. Keystroke dynamics represents one of the most popular and high-performance authentication methods among mobile behavioural biometrics \cite{STRAGAPEDE2023109089}.
\par
\begin{figure*}[t]
    \centering
     \includegraphics[trim={0cm 0cm 0 0cm}, width=0.9\linewidth]{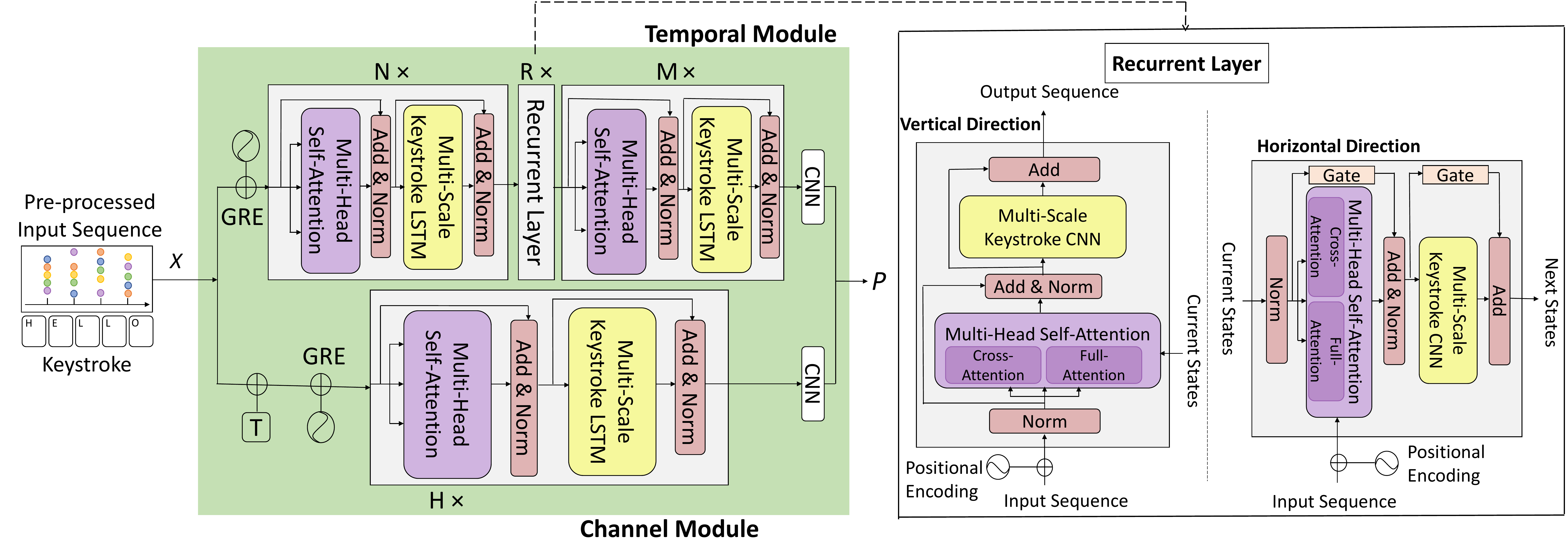}
    \caption{Graphical representation of TypeFormer, based on a Transformer architecture for biometrics keystroke free-text verification. T: Transposition operation; GRE: Gaussian Range Encoding; N, R, M, H: Number of layers of each of the modules; X: Pre-processed input sequence; P: Output feature embedding vector.}
    \label{fig:TransformerArch}
\end{figure*}

 In the present work, we propose a novel Transformer architecture, \textit{TypeFormer}, for mobile keystrokes dynamics for the purpose of user authentication. Transformers are recent Deep Learning (DL) networks, originally characterised by an encoder-decoder architecture \cite{vaswani2017attention}. Since their proposal, Transformers have been growing steadily due to their wide-ranging modelling abilities in several application fields such as computer vision, machine translation, reinforcement learning, time-series analysis for classification and prediction, etc. \cite{tay2020efficient}. 
In particular, in the present study we propose a Transformer network based on a two-branch (Temporal and Channel Modules) architecture with Long Short-Term Memory (LSTM) recurrent layers, Gaussian Range Encoding (GRE), a multi-head Self-Attention mechanism, and a Block-Recurrent Transformer layer (Fig. \ref{fig:TransformerArch}). TypeFormer is able to map slices of keystroke sequences into a feature embedding space where representations of sequences belonging to the same subject (intra-subject variability) are closer than those belonging to different subjects (inter-subject variability). TypeFormer is trained with the triplet loss function and the similarity of the feature embeddings is measured with Euclidean distance. %By replicating the experimental protocol and datasets of two recent related studies \cite{acien2021typenet, StragapedeMobile}, the proposed system is able to outperform them. 
\par In this way, while subjects type freely on their devices, TypeFormer might verify their identities passively by comparing and processing continuously acquired data samples with previously acquired and processed enrolment data (Fig. \ref{fig:workflow}).
\par In brief, the main contributions of the current work are: 
\begin{itemize}
    \item We propose \textit{TypeFormer}, a novel Transformer architecture for biometrics keystroke free-text verification, and provide an analysis of the different modules that compose the final architecture (Fig. \ref{fig:TransformerArch}).  %To the best of our knowledge, this is the first study that explores Transformers for keystroke biometrics.
    \item We perform an in-depth comparison with recent state-of-the-art keystroke verification systems based on LSTM Recurrent Neural Networks (RNN) and Transformers. By replicating the experimental protocol and adopting the same dataset \cite{palin2019people}, we outperform previous approaches \cite{acien2021typenet, StragapedeMobile} in terms of Equal Error Rate (EER), i.e., 3.25\% using only 5 enrolment sessions consisting in 50-keystroke sequences. As a result, we also reduce the traditional performance gap existing between mobile free-text and desktop fixed-text scenarios. Finally, we also analyse the behaviour of the model with different experimental configurations such as the length of the keystroke sequences and the amount of enrolment sessions.
    \item We include a cross-database evaluation of TypeFormer to assess the generalisation ability of the features extracted, showing that the proposed model is more robust in comparison with recent approaches such as TypeNet
 \cite{acien2021typenet}.   
    \item We make our experimental framework available to the research community, aiming to contribute to advancing the state of the art of keystroke biometrics\footnote{\texttt{\url{https://github.com/BiDAlab/TypeFormer}}}.

\end{itemize}

\par The remainder of the article is organised as follows: Sec. \ref{sec:Related_Works} describes key aspects of keystroke and Transformers. Then, Sec. \ref{sec:System_Description} presents the architecture of TypeFormer. The main characteristics of the databases considered are reported in Sec. \ref{sec:Database_Description}. In Sec. \ref{sec:Experimental_Protocol}, a detailed description of the experimental setup is reported. Sec. \ref{sec:Experimental_Results} contains the experimental results and the comparison with the state of the art. Finally, in Sec. \ref{sec:Conclusions_and_Future_Work} we sum up our contributions, and expose future research lines.

\section{Related Works}
\label{sec:Related_Works}

% TOUCHSCREEN behaviourAL MOBILE
%[sensor-based continuous authentication] 

%\cite{Sim2007}.
%\par Some early work in keystroke-based systems for authentication on mobile devices was carried out in \cite{Saevanee2008} by Saevanee \textit{et al.}. Using kNN, they obtained promising results in terms of accuracy, exploiting 10-digit sequences on a dataset collected involving 10 subjects. 
% \par Zahid \textit{et al.} \cite{Zahid2009}  studied mobile keystroke behaviour of 25 subjects, including features such as the hold time, error rate, and latency. The authors suggested a fuzzy classifier to account for the diffused features space and argued that presenting the classification task of keystroke behaviour as an optimisation problem benefits the robustness of the model when compared to similarity-based methods.

\subsection{Keystroke Biometrics}
\label{subsec:Keystroke_Biometrics}
Raw keystroke data generally consist in the timestamps of the actions of pressing and releasing a key, the key code typed, and additional features depending on the specific acquisition device such as the pressure and the area size of the finger. From the raw data, several features are commonly extracted:
\begin{itemize}
\item Latencies, i.e., the time intervals of press-to-press, press-to-release (which is also known as the \textit{hold time}), release-to-release, and release-to-press (\textit{fly time}) events.
\item Frequencies, such as the number of times per second a key is pressed or released. 
\item Error rates, related to the usage of backspaces or deletion options.
\item Screen coordinates (\textit{x}, \textit{y}) and their displacement, angles, velocity, acceleration, etc.
\end{itemize}

% Maiorana2021

%, adopted on computers before their application to smartphones \cite{Monrose1997}, where they typically achieve lower recognition performance \cite{acien2021typenet}. Nevertheless, in challenging mobile scenarios, keystroke dynamics usually outperforms other analogous behavioural traits such as swipe or tap gestures \cite{STRAGAPEDE202235}.

Moreover, a typical classification of the keystroke systems is based on the text format \cite{mondal2017study}: \textit{fixed text} (also known as \textit{text-dependent}), in which the sequences of the keys typed by the user are pre-determined, as in the case of login credentials, and \textit{free text} (\textit{text-independent}), in which the sequences of keys typed are arbitrary, as in the case of messages. The latter entails additional challenges in comparison to the former, i.e., the unstructured and sparse nature of the information captured, more frequent typing errors, and differences in between enrolment and verification sessions, leading to a higher intra-subject variability. The performance might also be affected if the same subject is able to speak different languages \cite{Abuhamad2021}. As a result, the performance reachable in the free-text scenario is usually worse than in the case of the fixed-text one \cite{acien2021typenet}. 
%Additionally, keystroke dynamics is one of the earliest behavioural authentication methods, adopted on desktop and laptop computers before their application to mobile devices, which entails additional challenges. 
%The mobile scenario has developed 
% In addition, focusing on keystroke biometrics on mobile scenarios, many challenges must be considered to develop robust authentication systems. 
\par Although biometric recognition based on keystroke has been investigated for over a  decade \cite{Maiorana2021, 9853506}, it can be still considered a biometric modality at the early stages, especially for mobile devices. In fact, before their application to mobile touchscreens, keystroke dynamics have been studied on the mechanical keyboards of desktop and laptop computers, for which, up to date, more in‐depth evaluations have been conducted and commercial applications have been proposed \cite{Maiorana2021}. In addition, mobile devices entail further challenges with respect to desktop ones, such as the unconstrained and non-stationary acquisition conditions, possibly due to the users' activity, body position, emotional state, etc. \cite{TEH2016210}. 
\par We describe next some of the key factors in the development and evaluation of a keystroke dynamics system:
\begin{itemize}[noitemsep]
    \item Authentication performance, quantified through popular metrics in the field of biometrics, such as EER, False Acceptance Rate (FAR), True Acceptance Rate (TAR), accuracy, Area Under the Curve (AUC), etc.
    \item Number of data subjects included in the database for development and evaluation of the technology.
    \item Amount of data required for each subject, i.e., number and duration of enrolment and verification sessions.
    \item Text format: fixed text, transcript or fully free text.
    \item Time interval between two acquisition sessions of the same subject, which can be a major source of variability due to \textit{biometric ageing}, as observed in other behavioural biometric modalities \cite{Tolosana_ageing}.
    \item  Information acquired, such as the timestamps of the actions of pressing and releasing a key, the key code typed, and additional features depending on the specific acquisition device such as the pressure.
    \item Instructions given to the subject during data acquisition which can can lead to a restricted acquisition environment.
    \item Other parameters such as the memory required to store and deploy the model, prediction time, etc.
\end{itemize}
A typical issue of the field of keystroke biometrics is the heterogeneity of databases, experimental protocols, and metrics. Therefore, a rigorous comparison between the different performance values is a difficult operation. %, not advisable if not under case-specific premises. 
To alleviate this aspect, Morales \textit{et al.} provided a common experimental framework for the fixed-text format by presenting the Keystroke Biometrics Ongoing Competition (KBOC) for user authentication using keystroke biometrics \cite{kboc}. 
%To this end, efforts towards a unification of the performance evaluation in the case of fixed-text scenario have been carried out by authors of the two public Aalto University databases, the one with desktop keyboard \cite{Dhakal2018} and the one collected on smartphone \cite{Palin_AaltoDBMobile19}, two of the widest and most complete keystroke databases, which we adopt in the current work (Sec. \ref{sec:Keystroke_Datasets}). 

\begin{table*}[!htbp]
\caption{Summary of different approaches presented in the literature for keystroke dynamics verification.}
%\scriptsize
\centering
\label{keystroke_table}
%\resizebox{21cm}{!}{ 
% \begin{tabular}{l c c c c c c}
% \vspace{.15cm}
\begin{adjustbox}{width=\textwidth,center}
\begin{tabular}{l c c c c c c c}
\hline
\textbf{Study} & \makecell{\textbf{Database}\\ \textbf{(Public)}} & \makecell{\textbf{Number of}\\ \textbf{Subjects}} & \textbf{Scenario} & \textbf{Classifier$^1$} & \textbf{Performance [\%]} & \makecell{\textbf{Text}\\ \textbf{Format}} & \makecell{\textbf{Data}\\\textbf{Amount}} \\
\hline
\hline
\makecell[l]{Monrose and Rubin \cite{Monrose1997a}\\ (1997)} & \makecell{Self-Collected\\ (\xmark)} & 42 & \textit{$\mathcal{D}$} & Weighted Euclidean dist. & \makecell{90.7 (Acc.) for Fixed Text\\23.0 (Acc.) for Free Text} & \makecell{Fixed,\\ Free} & Few sentences \\	
\hline
\makecell[l]{Gunetti and Picardi \cite{Gunetti2005} \\(2005)} & \makecell{Self-Collected\\ (\xmark)} & 205 & \textit{$\mathcal{D}$} & Different distance measures & $<$0.005 (FAR), $<$5 (FRR) & Free & 700-900 characters\\
\hline
\makecell[l]{Jiang \textit{et al.} \cite{Jiang2007} \\(2007)} & \makecell{Self-Collected\\ (\xmark)} & 58 & \textit{$\mathcal{D}$} & HMM & 2.54 (ERR) & Fixed & 20 strokes on average\\	
\hline
\makecell[l]{Saevanee \textit{et. al} \cite{Saevanee2008} \\(2008)} & \makecell{Self-Collected\\ (\xmark)} & 10 & \textit{$\mathcal{M}$} & kNN & 99.0 (Accuracy) & Fixed & 10-digit numbers\\
\hline
\makecell[l]{Killourhy and Maxion \cite{Killourhy2009} \\(2009)} &\makecell{CMU Database\\ (\cmark)} & 51 & \textit{$\mathcal{D}$} & \makecell{Manhattan dist., kNN, \\SVM, Mahalanobis, \\ NN, Euclidean dist.,\\ FL, \textit{k}-means} & \makecell{0.096 (EER) with\\ Manhattan dist.} & Fixed & 10 keystrokes\\
\hline
\makecell[l]{Zahid \textit{et al.} \cite{Zahid2009} \\(2009)} & \makecell{Self-Collected\\ (\xmark)} & 25 & \textit{$\mathcal{M}$} & FL, PSO & 2.07 (FAR), 1.73 (FRR) & Fixed & 250 keystrokes\\
\hline
\makecell[l]{Hwang \textit{et al.} \cite{Hwang2009} \\(2009)} & \makecell{Self-Collected\\ (\xmark)} & 25 & \textit{$\mathcal{M}$} & FF-MLP, RBFN, NN & 4 (EER) & Fixed & 4 digits \\
\hline
\makecell[l]{Giot \textit{et al.} \cite{Giot2011} \\(2011)} & \makecell{GREYC Web-Based\\ (\cmark) \cite{Giot2011}} & 100 & \textit{$\mathcal{D}$} & SVM & 15.28 (EER) & Fixed & 5 captures \\	
\hline
\makecell[l]{Balagani \textit{et al.} \cite{Balagani2011} \\(2011)} & \makecell{Self-Collected\\ (\xmark)} & 34 & \textit{$\mathcal{D}$} & SVM & \makecell{$<$1\\ (Average Error Rate)} & Free text & 500 keystrokes\\
\hline
\makecell[l]{Deng and Zhong \cite{Deng2013} \\(2013)} & \makecell{CMU Database\\ (\cmark) \cite{Killourhy2009}} & 51 & \textit{$\mathcal{D}$} & GMM, NN & 3.5-5.5 (EER) & \makecell{Fixed,\\ Free} & 1 sequence\\	
\hline
\makecell[l]{Ahmed \textit{et al.} \cite{Ahmed2013} \\(2013)} & \makecell{Self-Collected\\ (\cmark)} & 53 & \textit{$\mathcal{D}$} & Neural Network & \makecell{Controlled: 2.13 (EER,\\ 0 FAR, 5 FRR)\\ Uncontrolled: 2.46 (EER,\\ 0.01 FAR, 4.8 FRR)} & Free & 500 actions\\
\hline
\makecell[l]{Gascon \textit{et al.} \cite{gascon2014} \\(2014)} & \makecell{Self-Collected\\ (\xmark)}& 300 & \textit{$\mathcal{M}$} & SVM & 92 (TAR at 1\% FAR) & Free & 160 keystrokes\\
\hline
\makecell[l]{Alpar \cite{ALPAR2014213} \\(2014)} & \makecell{Self-Collected\\ (\xmark)}& 10 & \textit{$\mathcal{D}$} & \makecell{NN,\\ RGB histograms} & 90 (Acc.) & Fixed & 15 characters \\ 
\hline
\makecell[l]{Huang \textit{et al.} \cite{huang2015} \\(2015)} & \makecell{Clarkson I\\ (\cmark) \cite{clarkson1}} & 39 & \textit{$\mathcal{D}$} & Same as \cite{Gunetti2005} & \makecell{$\sim$1\\ (Impostor Pass Rate)} & Free & 1k-10k keystrokes \\
\hline
\makecell[l]{Morales \textit{et al.} \cite{kboc} \\(2016)} & \makecell{BiosecurID\\ (\cmark) \cite{fierrez2010biosecurid}} & 300 & \textit{$\mathcal{D}$} & Manhattan & 5.32 (EER) & Fixed & $\sim$25 keystrokes \\
\hline
\makecell[l]{Çeker and Upadhyaya \cite{ceker2016} \\(2016)} & \makecell{Clarkson I\\ (\cmark) \cite{clarkson1}} & 34 & \textit{$\mathcal{D}$} & SVM & $\sim$0 (EER) & Free & 500 keystrokes \\
\hline
\makecell[l]{Çeker and Upadhyaya \cite{ceker2017} \\(2017)} & \makecell{CMU Database (\cmark) \cite{Killourhy2009},\\ GREYC Keystroke (\cmark) \cite{giot2009},\\ GREYC Web-Based (\cmark) \cite{Giot2011}} & 267 & \textit{$\mathcal{D}$} & CNN & 2.02 (EER) & Free & Few keystrokes \\
\hline
\makecell[l]{Crawford \textit{et al.} \cite{crawford2017} \\(2017)} & \makecell{Self-Collected\\ (\xmark)} & 36 & \textit{$\mathcal{M}$} & Decision Tree & $>$93 (AUC) & Free & Few keystrokes \\
\hline
\makecell[l]{Kim \textit{et al.} \cite{Kim2018} \\(2018)} & \makecell{Self-Collected\\ (\xmark)} & 150 & \textit{$\mathcal{D}$} & 
\makecell{GDE, PWDE,\\ 1-SVM, \textit{k}-NN,\\ and \textit{k}-means} & \makecell{(EER: 0.44 for Korean,\\ 0.84 for English)} & Free & 100-1000 keystrokes \\
\hline
\makecell[l]{Murphy \textit{et al.} \cite{Murphy2017} \\(2017)} & \makecell{Clarkson II\\ (\cmark) \cite{Murphy2017}} & 103 & \textit{$\mathcal{D}$} & Same as \cite{Gunetti2005} & 2.17-10.7 (EER) & Free & 1000 keystrokes\\
\hline
\makecell[l]{Monaco \textit{et al.} \cite{Monaco2018} \\(2018)} & \makecell{CMU Database (\cmark) \cite{Killourhy2009}, (\cmark) \cite{monaco2},\\ (\cmark) \cite{monaco3}, (\cmark) \cite{monaco4}, (\cmark) \cite{monaco5}} & $\sim$50 & \textit{$\mathcal{D}$} & POHMM & \makecell{0.6-9 (EER),\\ 60.7-97.1 (Accuracy)} & \makecell{Fixed,\\ Free} & \makecell{0.12-55.18\\ events (on average)}\\	
\hline
\makecell[l]{Cilia \textit{et al.} \cite{Cilia2018} \\(2018)} & \makecell{Self-Collected\\ (\cmark)} & 24 & \textit{$\mathcal{M}$} & SVM & 0.44-3.93 (EER) & Fixed & Sentence based \\
\hline
\makecell[l]{Lu \textit{et al.} \cite{lu2019} \\(2020)} & \makecell{SUNY Buffalo (\cmark) \cite{buffalo},\\ Clarkson II (\cmark) \cite{Murphy2017}} & 75 & \textit{$\mathcal{D}$} & CNN + RNN & 2.67 (EER) & Free & 30 keystrokes \\
\hline
\makecell[l]{Kim \textit{et al.} \cite{kim2020} \\(2020)} & \makecell{Self-Collected\\ (\cmark)} & 50 & \textit{$\mathcal{M}$} & KS stat & $<$0.05 (EER) & Free & $\sim$200 keystrokes\\
\hline
\makecell[l]{Ayotte \textit{et al.} \cite{ayotte2020} \\(2020)} & \makecell{SUNY Buffalo (\cmark) \cite{buffalo},\\ Clarkson II (\cmark) \cite{Murphy2017}} & 101, 148 & \textit{$\mathcal{D}$} & RF & 7.8 (EER) & Free & 200 digraphs\\
\hline
\makecell[l]{Acien \textit{et al.} \cite{acien2021typenet} \\(2021)} & \makecell{Aalto Databases (\cmark) \cite{Dhakal2018, palin2019people},\\ SUNY Buffalo (\cmark) \cite{buffalo},\\ Clarkson II (\cmark) \cite{Murphy2017}} & 168K & \textit{$\mathcal{D, M}$} & RNN & \makecell{9.2 (EER) for \textit{$\mathcal{M}$},\\ 2.2 for \textit{$\mathcal{D}$}} & Free & 30-150 keystrokes \\
\hline
\makecell[l]{El-Kenawy \textit{et al.} \cite{math10162912} \\(2022)} & \makecell{RHU Dataset \cite{Rhu},\\ MEU-Mobile KSD Dataset \cite{7470816}} & 101, 148 & \textit{$\mathcal{M}$} & Bi-RNN & 99.02 (Acc.), 99.32 (Acc.) & Fixed & Few keystrokes \\
\hline
\makecell[l]{Stylios \textit{et al.} \cite{stylios2022} \\(2022)} & \makecell{Self-Collected\\ (\cmark)} & 39 & \textit{$\mathcal{M}$} & MLP & 97.18 (Acc.) & Fixed & $\sim$2 minutes sessions \\
\hline
\makecell[l]{Li \textit{et al.} \cite{Li2022} \\(2022)} & \makecell{SUNY Buffalo (\cmark) \cite{buffalo},\\ Clarkson II (\cmark) \cite{Murphy2017}} & 101, 148 & \textit{$\mathcal{D}$} & CNN + RNN & 97.68 (Acc.), 88.62 (Acc.) & Free & 50 keystrokes \\
\hline
\makecell[l]{Stragapede \textit{et al.} \cite{StragapedeMobile} \\(2022)} & Aalto Database \textit{$\mathcal{M}$} (\cmark) \cite{palin2019people} & 60K & \textit{$\mathcal{M}$} & Transformer & 3.84 (EER) & Free & 50 keystrokes \\
\hline
\makecell[l]{\textbf{TypeFormer}\\ \textbf{(2022)}} & \makecell{Aalto Databases (\cmark) \cite{Dhakal2018, palin2019people},\\ SUNY Buffalo (\cmark) \cite{buffalo},\\ Clarkson II (\cmark) \cite{Murphy2017}} & \textbf{60K} & \textbf{\textit{$\mathcal{D, M}$}} & \textbf{Transformer} & \textbf{3.25 (EER)} & \textbf{Free} & \textbf{30 - 100 keystrokes} \\
\hline
\multicolumn{8}{l}{\makecell[l]{$^1$Classifier Acronyms: \textit{HMM} = Hidden Markov Models, \textit{\textit{k}-NN} = k-Nearest Neighbours, \textit{SVM} = Support Vector Machine, \textit{NN} = Neural Network, \textit{FL} = Fuzzy Logic, \textit{PSO} = Particle Swarm\\ Optimisation, \textit{FF-MLP} =  Feed-Forward Multi-Layer Perceptron, \textit{RBFN} = Radial Basis Function Network, \textit{GMM} = Gaussian Mixture Model, \textit{CNN} = Convolutional NN, \textit{GDE} = Gaussian\\ Density Estimator,  \textit{PWDE} = Parzen Window Density Estimator, \textit{POHMM} = Partially Observable HMM, \textit{RNN} = Recurrent Neural Network, \textit{KS} = Kolmogorov-Smirnov, \textit{RF} = Random\\ Forest, \textit{Bi-RNN} = Bidirectional RNN, \textit{MLP} = Multi-Layer Perceptron.}} \\
\end{tabular}
\end{adjustbox}
% }
\end{table*}
% \end{landscape}

\subsection{Biometric Keystroke Verification}
\label{subsec:Biometric_Keystroke_Verification}
This section provides an overview of the key aspects of previous keystroke verification systems presented in the literature. The discussed studies are also reported in Table \ref{keystroke_table} in chronological order. %This section presents the evolution of the keystroke-based authentication systems proposed in the literature focusing on the proposed method. 
We consider systems developed in both desktop \textit{($\mathcal{D}$)}, and mobile \textit{($\mathcal{M}$)} scenarios. 
\subsubsection{Traditional Approaches}
\label{subsubsec:Traditional_Approaches}

%one of the major issues affecting the current state of the art is the lack of public databases. From the available databases stand out the  
% The latter shows that the analysis of the typing patterns performed on mobile devices has attracted far less interest than its counterpart on mechanical keyboards,  
%It also stands out HuMIdb \cite{Acien2020b} with   keystroke data on mobile devices.
% (systems), and Table \ref{keystroke_dataset_table} (databases). 
%During the usage of the device, when a key input is required (e.g., texting), the keystroke dynamics based authentication method validates the user since behavioural dynamics can be distinctive across users. Keystroke biometric systems are commonly placed into two categories: fixed-text, where the keystroke sequence typed by the user is prefixed, such as a username or password, and free-text, where the keystroke sequence is arbitrary, such as writing an email or transcribing a sentence with typing errors, and different between training and testing \cite{Acien2020a}. The performances of free-text algorithms are generally far from those reached in the fixed-text scenario, where the complexity and variability of the text entry contribute to intra-subject variations in behaviour, challenging the ability to recognise users \cite{Sim2007}. 
% The studies described are summarised in table \ref{keystroke_table}, and the used datasets in table \ref{keystroke_dataset_table}.

%%\iffalse[typenet]\fi 

\par In one of the earliest pioneering works on keystroke biometrics  \cite{Monrose1997a}, Monrose and Rubin proposed a free-text keystroke algorithm by using the mean latency and standard deviation of digraphs and computing the Euclidean distance between each test sequence and the reference profile\iffalse\textit{($\mathcal{D}$)}\fi. %Their results worsened from 90\% to 23\% of correct classification rates when they changed both user’s profiles and test samples from fixed-text to free-text. 
Gunetti and Picardi \cite{Gunetti2005} then extended the previous algorithm to n-graphs\iffalse\textit{($\mathcal{D}$)}\fi. 
%They calculated the duration of n-graphs common between training and testing and defined a distance function based on the duration and order of such n-graphs. Their results of 7.33\% classification error outperformed previous state of the art. Nevertheless, 
%Their algorithm needed long keystroke sequences (between 700 and 900 keys pressed) and many keystroke sequences (up to 14) to build the user’s profile, limiting the usability of that approach. 
More recently, due to their popularity, similar methods were used in \cite{huang2015} (2015) to study the effect of the data size on the performance of free-text keystroke\iffalse\textit{($\mathcal{D}$)}\fi, in \cite{crawford2017} (2017) to study how detecting the user’s position before authentication can significantly improve performance\iffalse\textit{($\mathcal{M}$)}\fi, and in \cite{Murphy2017} (2017) for benchmarking the large-scale database published, the Clarkson II database\iffalse\textit{($\mathcal{D}$)}\fi. The inclusion of time-related features such as rhythm and tempo was proposed in \cite{Hwang2009}\iffalse\textit{($\mathcal{M}$)}\fi. 
The Random Forest (RF) classifier was adopted in \cite{ayotte2020} to assess which are the most significant features of digraph-based algorithms (2020)\iffalse\textit{($\mathcal{D}$)}\fi.
\par A very popular method for keystroke biometrics is Support Vector Machine (SVM). Following previous findings, in \cite{Balagani2011} and \cite{ceker2016}, combinations of the existing digraphs method for feature extraction and a SVM classifier to authenticate users were proposed\iffalse\textit{($\mathcal{D}$)}\fi. 
SVM was also adopted in \cite{Giot2011}\iffalse\textit{($\mathcal{D}$)}\fi, and in \cite{gascon2014} in conjunction with mobile device background sensor data\iffalse\textit{($\mathcal{M}$)}\fi. Regardless of the classifier used, fusing keystroke dynamics with simultaneous movement sensor data included in mobile devices has proved to be very beneficial in terms of authentication results \cite{kim2020, STRAGAPEDE202235, STRAGAPEDE2023109089}\iffalse\textit{($\mathcal{M}$)}\fi. In a broad study (2018), Cilia \textit{et al.} \cite{Cilia2018} studied how differentiating typing modes (one or two hands) and user activity (standing or moving) during the development of a keystroke verification system based on SVM can improve the authentication performance significantly\iffalse\textit{($\mathcal{M}$)}\fi.
%This approach achieves almost 0\% error rate using samples containing 500 keystrokes. These results are very promising, even though it is evaluated using a small dataset with only 33 users. 
% {\color{red} Gascon \textit{et al.} \cite{gascon2014} 2014\iffalse\textit{($\mathcal{M}$)}\fi collected freely typed samples from over 300 participants and developed a system that achieved a True Acceptance Rate (TAR) of 92\% at 1\% False Acceptance Rate (FAR) (an EER of about 10\%). Their system utilised accelerometer, gyroscope, time, and orientation features. Each user typed an English pangram (sentence containing every letter of the alphabet) approximately 160 characters in length, and classification was performed by SVM.}

%found that models covering different typing modes and activities such as standing or moving and typing one handed or two handed, can be achieved with Equal Error Rate values as low as 0.44\%. The best results were achieved using a Least Squares SVM classifier with RBF kernel. They also explored and discussed any differences in classifying between typing activities, Digraphs vs. Trigraphs, feature importance, misclassification analyses and full typing sessions vs. sentence by sentence based classification. 

\par Among other classifiers, we mention Hidden Markov Models (HMM), used in \cite{Jiang2007} to exploit typing rhythms in keystroke dynamics\iffalse\textit{($\mathcal{D}$)}\fi, and then extended by Monaco \textit{et al.} \cite{Monaco2018} into Partially Observable Hidden Markov Models (POHMM)\iffalse\textit{($\mathcal{D}$)}\fi. %The difference from HMMs is that each hidden state is conditioned on an independent Markov chain. This algorithm is motivated by the idea that keystroke timings depend both on past events and the particular key that was pressed. Performance achieved using this approach in free-text is close to fixed-text, but it again requires several hundred keystrokes and has only been evaluated with a database containing less than 100 users. 
%\par Different studies have announced an extensive change in performance for comparable and similar algorithms, in light of the fact that most investigations utilised their very own dataset. To address this issue, 
With \textit{k}-Nearest Neighbour (\textit{k}-NN) \cite{Saevanee2008}\iffalse\textit{($\mathcal{M}$)}\fi, and fuzzy logic \cite{Zahid2009}\iffalse\textit{($\mathcal{M}$)}\fi, promising results have also been achieved in the early days of mobile keystroke biometrics. In the same epoch (2009), Killourhy and Maxion collected one of the first public databases of the field, the CMU keystroke dynamics database, and they carried out a benchmark evaluation with 14 different algorithms including Manhattan, Euclidean and Mahalanobis distances, \textit{k}-Nearest Neighbour, SVM (one-class), a neural network, fuzzy logic and \textit{k}-means \cite{Killourhy2009}\iffalse\textit{($\mathcal{D}$)}\fi. A similar benchmark study was conducted in \cite{Kim2018} on several algorithms such as Gaussian and Parzen Window Density Estimation, one-class SVM, \textit{k}-NN, and \textit{k}-means\iffalse\textit{($\mathcal{D}$)}\fi.

\subsubsection{Deep Learning Approaches}
\label{subsubsec:Deep_Learning_Approaches}

\par The advent of DL-based systems has not spared the field of keystroke biometrics, improving significantly the authentication performance, in particular in the more challenging free-text scenario.
In \cite{Deng2013} (2013), it was shown that a deep neural network was capable of outperforming other algorithms on the CMU keystroke dynamics database \cite{Killourhy2009}\iffalse\textit{($\mathcal{D}$)}\fi.
Approaches based on neural networks were also used for complementary tasks to improve the authentication performance, such as predicting the digraphs that are not present among the enrolment sessions by analysing the relation between the keystrokes \cite{Ahmed2013}\iffalse\textit{($\mathcal{D}$)}\fi. In \cite{ceker2017}, a Convolutional Neural Network (CNN) was introduced in combination with a Gaussian data augmentation technique for the fixed-text scenario\iffalse\textit{($\mathcal{D}$)}\fi, while in \cite{ALPAR2014213} a neural network was applied to RGB histograms obtained from fixed-text keystroke data. Moreover, Multi-Layer Perceptron (MLP) architectures have also been explored \cite{stylios2022} ($\mathcal{M}$).
\par In \cite{lu2019}, based on the observation that a RNN is a very suitable structure to learn from time-series \cite{8259229, 2020_TIFS_BioTouchPass2_Tolosana}, a combination of a convolutional and a recurrent network was proposed in order to extract higher level keystroke features on the SUNY Buffalo database \cite{buffalo} (2019)\iffalse\textit{($\mathcal{D}$)}\fi. The convolution process is performed before feeding the sequence to the recurrent network to characterise the keystroke sequence better. RNN variants are popular in keystroke biometrics, such as in \cite{math10162912}\iffalse\textit{($\mathcal{M}$)}\fi (birectional RNN), or in \cite{Li2022} ($\mathcal{M}$), in which keystroke sequences are arranged as an image-like matrix and then processed by a CNN combined with a Gated Recurrent Unit (GRU) network. In 2021, Acien \textit{et al.} presented TypeNet \cite{acien2021typenet}, a Siamese LSTM RNN for free-text keystroke biometrics. They considered the largest public databases to date, collected by researchers from the Aalto University, \cite{Dhakal2018}\iffalse\textit{($\mathcal{D}$)}\fi, and \cite{palin2019people}\iffalse\textit{($\mathcal{M}$)}\fi, with respectively around 168,000 and 68,000 subjects of free-text keystroke data divided into 15 acquisition sessions per subject. In their wide-ranging work, among other things, they achieved state-of-the-art authentication results at large scale in terms of EER (\%) while attempting to minimise the amount of data per subject required for enrolment. Following \cite{acien2021typenet}, in \cite{StragapedeMobile}, in 2022 we presented a preliminary attempt to use a Transformer architecture for keystroke biometrics, outperforming TypeNet in a specific experimental setup. We selected \cite{acien2021typenet} as a reference study for several reasons: (i) they adopt the largest mobile free-text keystroke databases available, the Aalto mobile keystroke database \cite{palin2019people}, (ii) their experimental protocol is publicly available on GitHub, allowing us to use the same sets of subjects and metrics, for development and evaluation, and (iii) they achieved state of the art results for free-text mobile keystroke biometrics. Consequently, references \iffalse\textit{($\mathcal{M}$)}\fi\cite{acien2021typenet} and \cite{StragapedeMobile} are particularly relevant to the current study as they use the same development and evaluation databases, and experimental protocol, allowing a direct comparison of the proposed systems (Sec. \ref{sec:Experimental_Results}).

\subsection{Introduction to Transformers}
\label{subsec:Transformers}
The first Transformer was proposed by Vaswani \textit{et al.} as a new encoder-decoder architecture \cite{vaswani2017attention}. Such model, later nicknamed the \textit{Vanilla} Transformer, is based purely on attention mechanisms, abandoning the idea of using convolutions or recurrence. The Vanilla Transformer was proposed for the task of machine translation, achieving remarkable results in comparison to existing systems in terms of quality of text translation and time consumption. In comparison with existing DL architectures such as CNNs or RNNs, the main advantages of the Transformer can be summarised as follows: \textit{(i)} all sequences are processed in parallel; \textit{(ii)} a Self-Attention mechanism is introduced to deal with long sequences; \textit{(iii)} the training is more efficient, modeling the whole sequences at once; \textit{(iv)} inspection of the whole sequences at once, without the need to summarise previous samples \cite{vaswani2017attention, xu2021autoformer, hutchins2022block}. 
\par Later, several variations of the original Transformer architecture have been proposed to overcome some of its drawbacks, and to deploy it in other application fields. In fact, its quadratic computational complexity and its considerable memory usage limited its application to longer time-series signals. To alleviate these aspects, the Two-stream Convolution Augmented Human Activity Transformer (THAT) was proposed by  Li \textit{et al.} for the task of Human Activity Recognition (HAR) \cite{li2021two}. Such architecture was designed based on the assumption that, similarly to images, time-series signals have information in two dimensions. Therefore, the model comprises two modules: \textit{(i)} the Temporal Module (extracting time features from unchanged data) and \textit{(ii)} the Channel Module (extracting channel features from transposed data). Then, the features extracted by each of the modules are concatenated for the prediction task. Another example of an interesting Transformer architecture variation is given by the Block-Recurrent Transformer, that has been recently introduced by Hutchins \textit{et al.} for the task of auto-regressive language modelling \cite{hutchins2022block}. In this approach, thanks to the recurrent on series-wise connections, all previous temporal information is retained. Furthermore, two attention mechanisms are applied at the same time (Full- and Cross-Attention).
\par In light of these and other adaptations, the popularity of Transformers increased in the last years due to the remarkable results obtained in other fields such as computer vision, reinforcement learning, time-series analysis for classification and prediction, biometrics, etc. \cite{tay2020efficient, delgado2022exploring}. 
A preliminary version of this work was published in \cite{StragapedeMobile} as the first application of Transformers to keystroke biometrics. This article significantly improves \cite{StragapedeMobile} in the following aspects: \textit{(i)} we propose a new Transformer architecture, TypeFormer, leading to an improvement of the authentication performance; \textit{(ii)} we provide a more extensive evaluation of the model, analysing the behaviour of the system with different experimental conditions such as the number of enrolment sessions and the length of the keystroke sequences; \textit{(iii)} we include a cross-database evaluation of TypeFormer considering other popular public databases, showing the ability of TypeFormer to generalise to other application scenarios; and \textit{(iv)} we provide an in-depth analysis of state-of-the-art keystroke verification systems, remarking key aspects such as the scenario (fixed or free text) and database considered, classifier, and performance.

\section{Proposed System: TypeFormer}
\label{sec:System_Description}

This section contains a detailed description of all aspects of the proposed keystroke verification system.

\subsection{Feature Extraction}
\label{subsec:Feature_Extraction}
\begin{figure}[tp]
    \centering
     \includegraphics[trim={0cm 0cm 0 0cm}, width=0.8\linewidth]{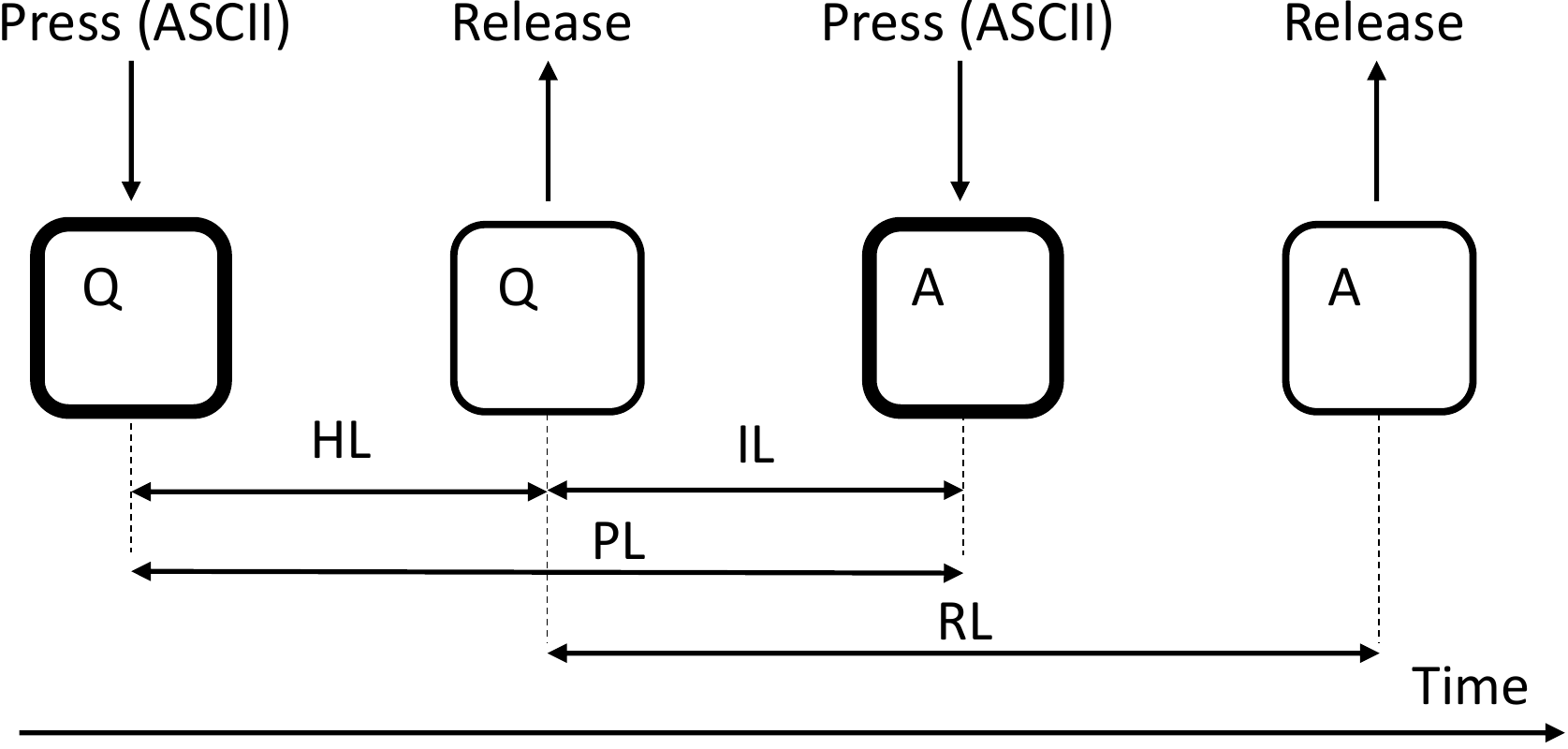}
    \caption{Example of the keystroke features extracted from the Aalto mobile keystroke database \cite{palin2019people}. HL: Hold Latency; IL: Inter-key Latency; PL: Press Latency; RL: Release Latency; ASCII: Key Pressed.}
    \label{fig:keystrokefeatures}
\end{figure}

%pre-processed following the approach described in \cite{acien2021typenet}
The raw keystroke information available consists essentially in the timestamp of the event of pressing (finger down) and releasing (finger up) a key, together with the ASCII code typed. Such data are processed to extract a set of 5 features per character typed:
\begin{center}
[\textit{hold latency}, \textit{inter-key latency}, \textit{press latency}, \textit{release latency}, \textit{key pressed}]
\end{center}
The above-mentioned features are shown in Fig. \ref{fig:keystrokefeatures}. Due to the fact that the length of the free-text sequences is not fixed, they are sliced or zero-padded to produce a fixed-size input, ($L = 30, 50, 70, 100$), depending on the specific experiment (see Sec. \ref{sec:Experimental_Protocol}). The ASCII code (key pressed) is normalised in the range $[0,1]$.

%In a free-text scenario, it would be preferable to avoid relying on the key pressed in order to disclose the content of the message for privacy-related purposes. However, in light the acquisition scenario of the considered database, in which the sentences are provided to the user, we opted to maintain such information, to provide a fair benchmark of our proposed system with the compared state-of-the-art system \cite{acien2021typenet}.

\subsection{TypeFormer Architecture}
\label{subsec:Transformer_Architecture}

% \par Fig. \ref{fig:TransformerArch} provides a graphical representation of the proposed mobile authentication system based on Transformers \cite{vaswani2017attention}. 
% The architecture presented in this article is an improvement of the previous work presented in \cite{StragapedeMobile}.
Following the same idea presented in \cite{li2021two}, TypeFormer contains two modules, each of them in a specific branch, to which the pre-processed Transfomer input sequences $X$ (Sec. \ref{subsec:Feature_Extraction}) are fed (please, see Fig. \ref{fig:TransformerArch} for a better understanding): a Temporal Module (temporal-over-channel features), and a Channel Module (channel-over-temporal features).
In both channels, $X$ is modelled using a GRE to preserve the information position. The output sequence is defined by an $L_{1}$ normalised vector representing the Probability Density Function (PDF) of the Gaussian distributions $G$. Moreover, the final GRE is calculated by a weighted multiplication over several ranges, containing the behaviour of each of the samples in a different scenario. 
\par The Temporal Module contains three ordered sets of layers. Each of the sets of layers is composed respectively by $N$, $R$, and $M$ layers. The $N$ and $M$ layers are identical, and made of two sub-layers: a multi-head Self-Attention mechanism, and a multi-scale keystroke LSTM RNN layer. The multi-head Self-Attention mechanism connect the samples among the whole sequence obtaining long-range dependencies. The mechanism applies a weighted sum of the different values $V$ over the different queries $Q$ and the matching keys $K$. The output of the Self-Attention sub-layer is the result of applying the attention mechanism to $F$ independent heads. Then, the multi-scale keystroke LSTM RNN layer is activated by ReLU functions. Each of the scales contains a unique kernel. Following each sub-layer, a residual connection and a layer normalisation are included (\textit{Add \& Norm} in Fig. \ref{fig:TransformerArch}). 
\par Between the $N$ and $M$ layers, $R$ recurrent layers are included (graphically represented in detail on the right side of Fig. \ref{fig:TransformerArch}). The structure of such layers is based on the Block-Recurrent Transformer architecture presented in \cite{hutchins2022block}. Initially, the input sequence is shaped by a positional encoding. Then, a recurrent form of attention is introduced in the vertical and horizontal directions, based on two sub-layers in each of the directions: \textit{(i)} a multi-head Self-Attention mechanism, which applies Full-Attention to the sequences to obtain the matching values $V$ and keys $K$, and Cross-Attention to the current states (initialised to 0) to extract the queries $Q$ (replicated in $F$ independent heads); \textit{(ii)} a multi-scale keystroke CNN network, which comprises a CNN with ReLU activations and unique kernels for each of the scales. Every sub-layer is preceded by a layer normalisation and followed by a residual connection (\textit{Add \& Norm}). While the multi-scale keystroke CNN network remains unchanged, the multi-head Self-Attention mechanism applies Cross-Attention to the sequences to obtain the matching queries $Q$, and Full-Attention to the current states to extract the keys $K$ and the values $V$ (such mechanism is replicated in $F$ independent heads). Furthermore, the residual connections are replaced by forget gates, altering the current states.
\par The Channel Module input sequence $X$ is transposed and modelled by the GRE. Then, $H$ layers (analogous to the $N$ and $M$ layers of the Temporal Module) are included, followed by a residual connection and a layer normalisation (\textit{Add \& Norm}). %Each layer consist of two sub-layers: \textit{(i)} a multi-head Self-Attention mechanism, and \textit{(ii)} a multi-scale keystroke LSTM. %Similar to the Temporal Module, each sub-layer is followed by a residual connection and a layer normalisation (\textit{Add \& Norm}).
\par Subsequently, each of the Modules is followed by a convolutional layer, after which the similarity of the output features are concatenated into an output vector $P$ and fed into a sigmoid layer. Finally, for the authentication task considered in the present study, the output feature embedding vectors are compared using the Euclidean distance. %It is important to highlight that the architecture configuration and model hyper-parameters of the proposed Transformer have been adapted to free-text keystroke authentication systems on mobile devices. 
The specific details of the hyper-parameter implementation for the proposed Transformer are described in Sec. \ref{subsec:Transformer_Hyperparameters}.

%\begin{figure}[tp]
%    \centering
%     \includegraphics[trim={0cm 0cm 0 0cm}, width=0.7\linewidth]{Figures/GaussianRangeEncoding2.pdf}
%    \caption{Graphical representation of the Gaussian range encoding. PDF: Probability Density Function.}
%    \label{fig:GaussianRangeEncoding}
%\end{figure}

\section{Databases Description}
\label{sec:Database_Description}

\subsection{Development Database}
\label{subsec:The_Development_Dataset}
The Aalto mobile keystroke database is a large-scale database for mobile keystroke biometrics involving around 260,000 subjects \cite{palin2019people}. In this work we have selected all subjects that completed at least 15 acquisition sessions, reducing the number of subjects to 62,454.
The raw data available in the Aalto mobile keystroke database consist in the timestamps of the key press (finger down) and key release (finger up) gestures with a 1 ms-resolution. The data was captured through a mobile web application in an unsupervised way. Subjects were asked to read, memorise, and type in their smartphone English sentences that were randomly selected from a set of 1,525 sentences obtained from the Enron mobile mail \cite{10.1145/2037373.2037418} and the Gigaword Newswire corpora \cite{gigaword}. Therefore, the text format adopted is free-text, with sentences containing at least 3 words or 70 characters. Moreover, the volunteers were asked to type as fast and accurately as possible. Concerning the volunteers, they were selected from 163 countries, approximately 68\% of the subjects involved were English native speakers, and around 31\% of them took a typing course.

\subsection{Evaluation Databases}
\label{subsec:Other_Evaluation_Datasets}
Apart from the Aalto mobile keystroke database, three other databases are considered in the current work to carry out a cross-database evaluation and to assess the generalisation ability of the features extracted by TypeFormer. They are:
\begin{itemize}
    \item The Aalto desktop keystroke database was presented in \cite{Dhakal2018}. The collection of the desktop database took place earlier than its mobile counterpart, but the acquisition settings were similar across the two, apart from the use of a mechanical keyboard. The desktop database comprises over 168,000 participants with at least 15 sentences. 72\% of the participants from the desktop database took a typing course, 218 countries were involved, and 85\% of the them are English native speakers.
    \item The Clarkson II database \cite{Murphy2017} involves 103 subjects. The acquisition took place in a desktop environment in a 2.5-year span in a completely unsupervised scenario and totally free-text. There were no separate acquisition sessions, therefore in order to obtain the enrolment and verification sessions, each of the data sequences was split in shorter sequences. To obtain a similar testbench as the Aalto databases, in our evaluation we include only the subjects with at least 15 keystroke sequences.
    \item The Buffalo database \cite{buffalo} contains data from 148 subjects, divided into 3 separate acquisition sessions. The data were collected over a 28-day time span from mechanical keyboards (desktop environment). The Buffalo database is split into two tasks (text transcription and completely uncontrolled free text).
\end{itemize}

\section{Experimental Protocol}
\label{sec:Experimental_Protocol}
\subsection{TypeFormer Hyperparameters}
\label{subsec:Transformer_Hyperparameters}

The best configuration found in terms of the hyperparameters of the proposed Transformer is described below. The Gaussian range encodings contain $G = 20$ Gaussian distributions. The Temporal Module comprises $N = 9$, $R = 2$, and $M = 1$ layers with $F = 10$ heads each, while the Channel Module $H = 1$ layer with $F = 5$ heads. In both modules the multi-scale keystroke LSTM contains 3 recurrent layers with kernel sizes 1, 3, and 5, respectively. Each of them comprise $D$ units and ReLU activation functions, followed by dropout layers with a rate of 0.1. The multi-scale keystroke CNN networks of the $R$ recurrent layers contain $D$ units each (where $D$ corresponds to the keystroke sequence length $L$), ReLU activation functions, and kernel sizes 1, 3, and 5, respectively, followed by dropout layers with a rate of 0.1. Subsequent to the Temporal and Channel Modules, 2 convolutional layers are included with $D$ units, ReLU activation functions, and kernel sizes 128 and 32 respectively. Each of the convolutional layers are followed by dropout layers with a rate of 0.5. Finally, a max-pooling layer followed by a linear layer with sigmoid activation function are included. The final output vector contains $S = 64$ features.

\subsection{Model Development}
\label{subsec:Model_Development}
In order to perform a fair comparison across different DL architectures, in the current work we replicate the public experimental protocol presented by Acien \textit{et al.} in \cite{acien2021typenet}. Specifically, data belonging to the same non-overlapping 30,000 and 400 subjects have been used respectively for the purpose of training and validation. Each subject data are organised into 15 acquisition sessions. The triplet loss function is employed for the training, and a margin of $\alpha = 1.0$ was set on top of the Euclidean distance for each of the pair combinations in the triplet. Additionally, the Adam optimiser with a learning rate of 0.001 is used. The Transformer is trained for 1,000 epochs, considering roughly 30,000 triplets per epoch, arranged into 1024-sequence-sized batches. The triplets are formed by sampling subjects randomly and with uniform distribution across the training set. At the end of each training epoch, the model performance is quantified in terms of EER, and according to such metric the best model is selected to be tested on the final evaluation subset. TypeFormer is implemented in \texttt{PyTorch}.

\subsection{Model Evaluation}
\label{subsec:Model_Evaluation}
We describe next the experiments considered in the present study to validate the proposed TypeFormer. In all of them, different subjects are used for training and evaluating the keystroke verification model.

\subsubsection{Experiment 1: Intra-Database Evaluation}
\label{subsubsec:Experiment_1}
The first experiment analyses the performance of TypeFormer over an evaluation set of $U =  1,000$ unseen subjects obtained from the same database considered in training. At the end of each of the training epochs, the best model is selected using a separate validation subset. We follow the same protocol as \cite{acien2021typenet}, considering $E$ enrolment sessions per subject. The genuine and impostor score distributions are subject-specific. For each subject, genuine scores are obtained comparing the enrolment sessions ($E$) with  5 verification sessions. The Euclidean distances are computed for each of the verification sessions with each of the $E$ enrolment sessions, and then values are averaged over the enrolment sessions. Therefore, for each subject there are 5 genuine scores, one for each verification session. Concerning the impostor score distribution, for every other subject in the evaluation set, the averaged Euclidean distance value is obtained considering 1 verification session and the above-mentioned 5 enrolment sessions. Consequently, for each subject, there are 999 impostor scores. Based on such distributions, the EER score is calculated per subject, and all EER values are averaged across the entire evaluation set. The number of enrolment sessions is variable ($E = 1, 2, 5, 7, 10$) in order to assess the performance adaptation of the system to reduced availability of enrolment data. Additionally, also the experiments are repeated changing the input sequence length, $L = 30, 50, 70, 100$, to evaluate the optimal keystroke sequence length.

\subsubsection{Experiment 2: Cross-Database Evaluation}
\label{subsubsec:Experiment_2}
A key aspect of machine learning is the generalisation ability of the system, in other words, its ability to work well with different databases from those used during the development stage. Such assessment is known as \textit{cross-database} evaluation. Designing a model capable of extracting robust features is a challenging task. 
In this experiment, once again, we take \cite{acien2021typenet} as the reference study, and replicate their protocol to compare TypeFormer with the state of the art. Therefore, the Aalto desktop keystroke database \cite{Dhakal2018}, the Clarkson II \cite{Murphy2017}, and the SUNY Buffalo \cite{buffalo} databases are considered. 
%In this experiment, we employ the best model obtained from Experiment 1 (Sec. \ref{subsubsec:Experiment_1}). 
Such databases were selected as they are popular in the literature (see Table \ref{keystroke_table}), and publicly available. For consistency, we consider $E = 5$ enrolment sessions, $L = 50$ keystrokes per session, and $U =  1,000$ test subjects for the Aalto desktop database. Regarding the Clarkson II database, $U =  91$ subjects are considered (the number of subjects for which we could extract at least 15 sessions of 150 keys), $E = 5$ enrolment sessions per subject, $L = 50$ keystrokes per sequence. For the SUNY Buffalo database, $U =  147$, $E = 2$ enrolment sessions per subject (as there are only three sessions per subject), and $L = 50$ keystrokes per sequence.

\section{Experimental Results}
\label{sec:Experimental_Results}

\begin{table}[t]
\small
\centering
\caption{Experimental results of the different modules implemented in the development of TypeFormer, in comparison with the Vanilla Transformer \cite{vaswani2017attention}\\($E$ is the number of enrolment sessions).}
% \vspace{.15cm}
\footnotesize
\begin{tabular}{|c|c|c|c|c|c|}
\hline
\textbf{System} & \textbf{E = 1} & \textbf{E = 2} & \textbf{E = 5} & \textbf{E = 7} & \textbf{E = 10} \\ \hline
\makecell{Vanilla\\ Transformer \cite{vaswani2017attention}} & 10.28 & 8.56 & 7.41 & 6.95 & 6.61 \\ \hline
\makecell{Temporal Branch\\ w/o Rec. Layer} & 8.15 & 6.43 & 5.12 & 4.73 & 4.29 \\ \hline
\makecell{Temporal Branch\\ w/ ~Rec. Layer} & 7.12 & 5.49 & 3.94 & 3.63 & 3.15 \\ \hline
\makecell{Channel Branch} & 17.29 & 15.50 & 13.54 & 13.07 & 12.55 \\ \hline
\makecell{\textbf{TypeFormer} (Temp.\\ + Channel Branch\\ w/ Rec. Layer)} & \textbf{6.17} & \textbf{4.57} & \textbf{3.25} & \textbf{2.86} & \textbf{2.54} \\ \hline
\end{tabular}
\label{table:DifferentArch}
\end{table}

\begin{table*}[t]
    \small
    \centering
    \caption{Intra-database evaluation: System performance results in terms of EER for the final evaluation dataset of the Aalto mobile database.}
    % \vspace{.15cm}
    \begin{tabular}{{|c|c|c|c|c|c|c|}}
    \hline
    \multirow{2}{*}{\makecell{\textbf{Sequence}\\ \textbf{Length $L$}}} & 
    \multirow{2}{*}{\makecell{\textbf{Model}}} & 
    \multicolumn{5}{c|}{\makecell{\textbf{Number of Enrolment Sessions $E$}}} \\
    \cline{3-7}
     & & \textbf{1}\hfill
 & \textbf{2} & \textbf{5} & \textbf{7} & \textbf{10} \\
    \hline
 \multirow{2}{*}{\textbf{30}} & \multirow{1}{*}{Acien \textit{et al.} \cite{acien2021typenet}} &  14.20\ \ & 12.50\ \  & 11.30\ \ & 10.90\ \ & 10.50 \\
    \cline{2-7}
     & \textbf{TypeFormer} & \textbf{9.48} & \textbf{7.48} & \textbf{5.78} & \textbf{5.40} & \textbf{4.94} \\
    \hline
    \multirow{3}{*}{\textbf{50}} & \multirow{1}{*}{Acien \textit{et al.} \cite{acien2021typenet}} &  12.60 & 10.70 & 9.20 & 8.50 & 8.00 \\
     \cline{2-7}
     & Preliminary Transformer \cite{StragapedeMobile} &  6.99 & - & 3.84 & - & 3.15 \\
    \cline{2-7}
     & \textbf{TypeFormer} &  \textbf{6.17} & \textbf{4.57} & \textbf{3.25} & \textbf{2.86} & \textbf{2.54} \\
    \hline
    \multirow{2}{*}{\textbf{70}} & \multirow{1}{*}{Acien \textit{et al.} \cite{acien2021typenet}} &  11.30 & 9.50 & 7.80 & 7.20 & 6.80 \\
    \cline{2-7}
     & \textbf{TypeFormer} & \textbf{6.44} & \textbf{5.08} & \textbf{3.72} & \textbf{3.30} & \textbf{2.96} \\
     \hline
     \multirow{2}{*}{\textbf{100}} & \multirow{1}{*}{Acien \textit{et al.} \cite{acien2021typenet}} &  10.70 & 8.90 & 7.30 & 6.60 & 6.30 \\
    \cline{2-7}
     & \textbf{TypeFormer} & \textbf{8.00} & \textbf{6.29} & \textbf{4.79} & \textbf{4.40} & \textbf{3.90}\\
    \hline
\end{tabular}
    \label{table:TransformersComparisonTypeNet}
\end{table*}

\subsection{Experiment 1: Intra-Database Evaluation}
Starting from the initial Vanilla Transformer proposed in \cite{vaswani2017attention}, to validate each part of final proposed system, Table \ref{table:DifferentArch} presents the experimental results of the different modules implemented in the development of TypeFormer. The results are obtained on the final evaluation dataset of the Aalto mobile database. This analysis is carried out by considering a variable number of enrolment sessions $E = 1, 2, 5, 7, 10$ along the columns, and sequence length $L = 50$. Although the Vanilla Transformer is solely based on attention mechanisms, it shows the effectiveness of the Transformer architecture in modelling keystroke sequences. 
First, this architecture is modified by including the Gaussian Range Encoding (instead of the Positional Encoding originally used in the Vanilla Transformer). Then, the Point-Wise Feed-Forward Networks of the Vanilla Transformer are changed with LSTM recurrent layers (Temporal w/o Rec. Layer). By doing so, we obtain an improvement for all considered amounts of enrolment sessions, and the recognition performance in terms of EER is improved on average by a 28.70\%. Following \cite{hutchins2022block}, a Block-Recurrent Transformer layer is introduced in the Temporal branch in the case of the Temporal with Recurrent Layer configuration. This further reduces the EER by a 20.03\% (Temporal w/ Rec. Layer).
Then, we considered transposed input sequences in the Channel Branch configuration...
Finally, we considered the combination of the Temporal with Recurrent Layer and Channel Branch configurations, corresponding to the final TypeFormer architecture.

\begin{figure}[b!]
    \centering
     \includegraphics[trim={0cm 0cm 0 0cm}, width=.92\linewidth]{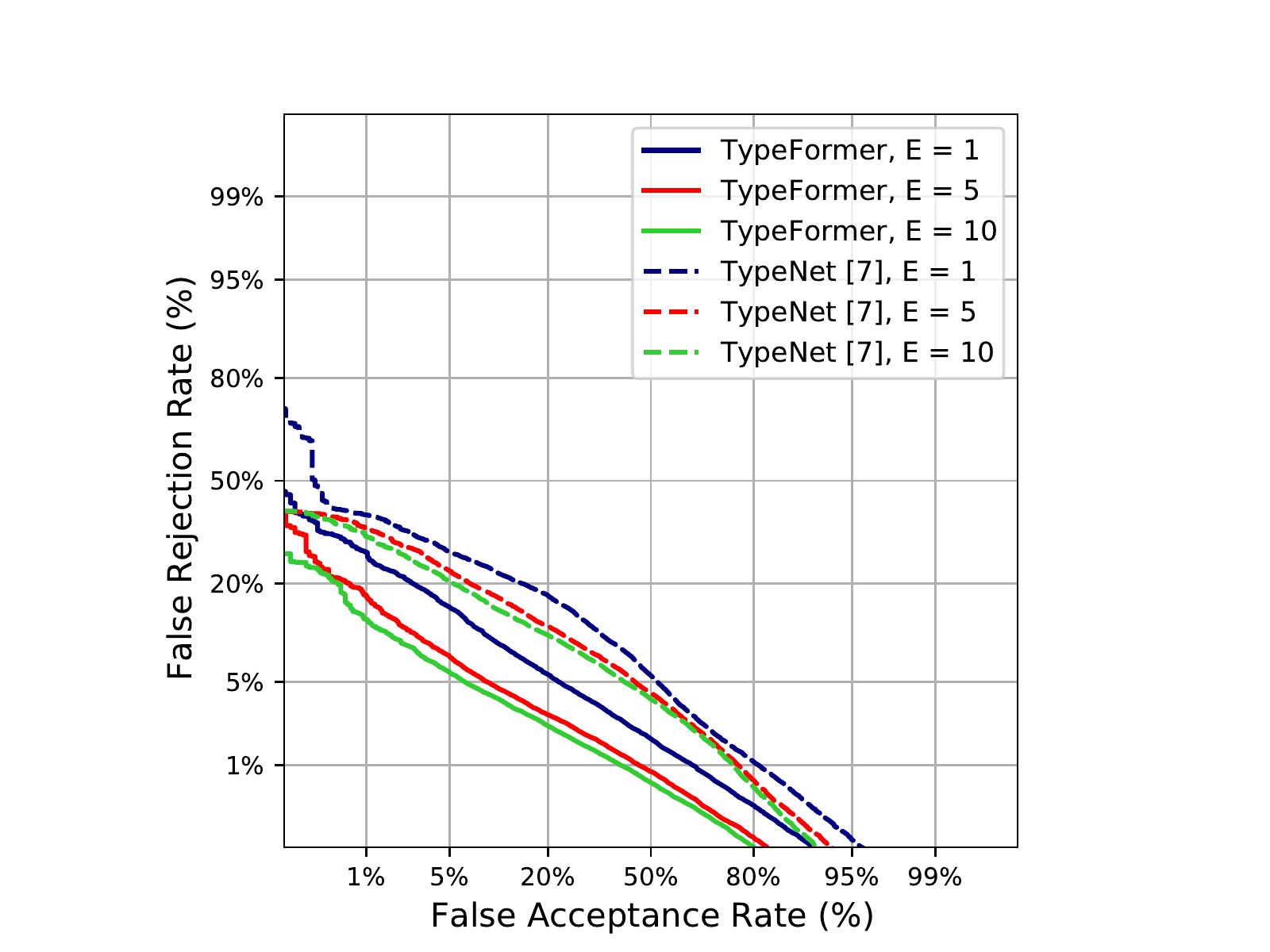}
    \caption{DET curves comparing the performance of TypeFormer with TypeNet (\cite{acien2021typenet}) for keystroke sequences of length $L = 50$. \textit{E} corresponds to the number of enrolment sessions considered.%The reported EERs (\%) are for the global threshold.
    }
    \label{fig:dets}
% \vspace{-0.5cm}
\end{figure}

Table \ref{table:TransformersComparisonTypeNet} shows the results achieved by TypeFormer considering different the sequence lengths $L$. 
In addition, to provide a better comparison of TypeFormer with recent state-of-the-art keystroke biometric systems, we include the results achieved by TypeNet in \cite{acien2021typenet}, and our preliminary study \cite{StragapedeMobile} on the same dataset as in the previous Table \ref{table:DifferentArch}.
%It is important to point out that such studies were chosen since \textit{(i)} they consider the largest public mobile free-text keystroke database to date, the Aalto mobile keystroke database \cite{palin2019people}, \textit{(ii)} the experimental protocol is reported on GitHub\footnote{\texttt{\url{https://github.com/BiDAlab/TypeNet}}}, making it possible to follow it rigorously, and \textit{(iii)} they achieved state-of-the-art results with a recent DL approach.
%To assess the ability of TypeFormer to work with different amounts of data, several different keystroke input sequence lengths are considered ($L = 30, 50, 70, 100$), as well as the number of enrolment sessions ($E = 1, 2, 5, 7, 10$).
In general, in Table \ref{table:TransformersComparisonTypeNet}) we can see that in all cases TypeFormer outperforms previous approaches over the same evaluation set of 1,000 subjects. In particular, the performance improvement of TypeFormer averaged over all cases in the table ($E = 1, 2, 5, 7, 10$ and $L = 30, 50, 70, 100$) consists in 47.3\% in relative terms with respect to TypeNet \cite{acien2021typenet}, an LSTM RNN-based system. To provide a graphical representation of the differences in the performance of the compared systems. Fig. \ref{fig:dets} reports the Detection Error Trade-off (DET) curves computed for the different number of enrolment sessions available ($L = 50$). The graph shows that our proposed approach outperforms the LSTM RNN of TypeNet in all cases, i.e., $E=1$ (TypeFormer) enrolment session vs. $E=10$ (TypeNet). %The difference is less among the two Transformer architectures, however the modifications introduced within TypeFormer lead to better performance. 
This shows the ability of TypeFormer to model keystroke dynamics. 
\par Additionally, considering only the results of Table \ref{table:TransformersComparisonTypeNet} obtained by TypeFormer, it is possible to observe that in all cases the EER values decrease as the number of enrolment sessions $E$ increases. 
Such trend is predictable and consistent for all sequence lengths $L$. Also, the rate of improvement is higher going from $E = 1$ to $E = 5$ sessions (relative improvement of almost 50\% going from 6.17\% to 3.25\% EER for $L = 50$) than from $E = 5$ to $E = 10$ (relative improvement of around 20\% going from 3.25\% to 2.54\% EER for $L = 50$).

\par Similarly, by carrying out an analogous analysis along the rows, it is noticeable that increasing the input sequence length $L$ from 30 to 50, there is a significant improvement (42.64\% in relative terms on average over all considered enrolment session amounts $E$) in terms of EER. Nevertheless, such trend is reversed when increasing the sequence length $L$ to 70 or 100 (respectively a performance degradation of 12.38\%, and 28.38\% in relative terms on average over all considered enrolment session amounts $E$), leading to the conclusion that the optimal sequence length must be around 50. This could be due to the fact that the zero-padding operation carried out to equalise the length of different keystroke sequences is not beneficial for the Transformer-based architecture that rely on an attention mechanism, that can perhaps be optimised. In case of the RNN-based reference system \cite{acien2021typenet}, the longer the input sequences, the better the results, showing the beneficial effects of the masking layer included in their network.
\par Lastly, Table \ref{table:TransformersComparisonSOTA} presents a comparison of the proposed TypeFormer with other systems presented in the literature that were not originally evaluated according to the protocol adopted in this work \cite{palin2019people}: digraphs and SVM \cite{ceker2016}, POHMMs \cite{monaco2018partially}, and a combination of RNNs and CNNs \cite{lu2019}. 
The evaluation of the different system takes place on the same set of 1,000 subjects considering $E = 5$ and $L = 50$.
TypeFormer shows the best performance, with EER absolute improvements of 37.15\% (POHMM \cite{monaco2018partially}), 32.45\% (Diagraphs \cite{ceker2016}), 8.95\% (CNN + RNN \cite{lu2019}), 5.95\% (TypeNet \cite{acien2021typenet}), and 0.59\% (our preliminary Transformer architecture \cite{StragapedeMobile}). Such results show the potential of TypeFormer and Transformer-based architectures in the challenging free-text mobile scenario.

\begin{table}[t!]
\small
\centering
\caption{Comparison of the performance achieved by the proposed TypeFormer with related systems that followed different experimental protocols in the studies in which they were originally proposed ($E$ = number of enrolment sessions = 5, $L$ = number of enrolment sessions considered = 50).}
% \vspace{.15cm}
\begin{tabular}{|c|c|}
\hline
\textbf{System}               & \textbf{EER {(}\%{)}} \\ \hline
POHMM \cite{monaco2018partially}           & 40.40                   \\ \hline
Digraphs \cite{ceker2016}        & 29.20                   \\ \hline
CNN+RNN  \cite{lu2019}            & 12.20                   \\ \hline
TypeNet   \cite{acien2021typenet}           & 9.20                    \\ 
\hline
Preliminary Transformer   \cite{StragapedeMobile}           & 3.84                    \\ \hline
\textbf{TypeFormer} & \textbf{3.25}                    \\ \hline
\end{tabular}
\label{table:TransformersComparisonSOTA}
\end{table}

\begin{table}[t!]
\small
\centering
\caption{Cross-Database Evaluation: EER (\%) achieved by TypeFormer in comparison with TypeNet \cite{acien2021typenet}. The databases considered are Aalto Mobile (development set) \cite{palin2019people}, Aalto Desktop \cite{Dhakal2018}, Clarkson II \cite{Murphy2017}, and SUNY Buffalo (free-text and transcripted text) \cite{buffalo} (all in the desktop scenario). *Experiments using all the available data per subject.}
% \vspace{.15cm}
\begin{tabular}{|c|c|c|}
\hline
\makecell{\textbf{Evaluation} \textbf{Database}}  & \makecell{\textbf{Acien \textit{et al.}} \textbf{\cite{acien2021typenet}}} & \makecell{\textbf{TypeFormer}} \\ 
\hline
Aalto Mobile & 9.20 & 3.25 \\ 
\hline
Aalto Desktop & 21.40 & 15.02 \\ 
\hline
Clarkson II & 36.60 & 27.83 \\ 
\hline
Clarkson II* & 33.00 & 25.34 \\ 
\hline
SUNY Buffalo (Free) & 33.20 & 22.39 \\ 
\hline
SUNY Buffalo (Transcript) & 32.80 & 23.40 \\ 
\hline
\end{tabular}
\label{table:OtherDatabases}
\end{table}

\subsection{Experiment 2: Cross-Database Evaluation}
Table \ref{table:OtherDatabases} shows the results obtained by deploying TypeFormer ($E = 5$, $L = 50$) on different databases and keystroke scenarios not considered during the development of the model (Aalto mobile keystroke database). This experiment is useful to assess the generality and robustness of the features extracted by TypeFormer. In general, we can observe that there is a significant performance degradation when considering different databases. Consequently, this aspect should not be underestimated for real-life applications. It is important to highlight that we have not considered any fine-tuning strategy of the model. Nevertheless, the proposed TypeFormer is able to mitigate significantly such effect in comparison to \cite{acien2021typenet}, reaching an absolute improvement of 8.60\% EER on average on the considered cross-database evaluation cases.

\begin{figure}[b!]
    \centering
     \includegraphics[trim={0cm 0cm 0 0cm}, width=0.92\linewidth]{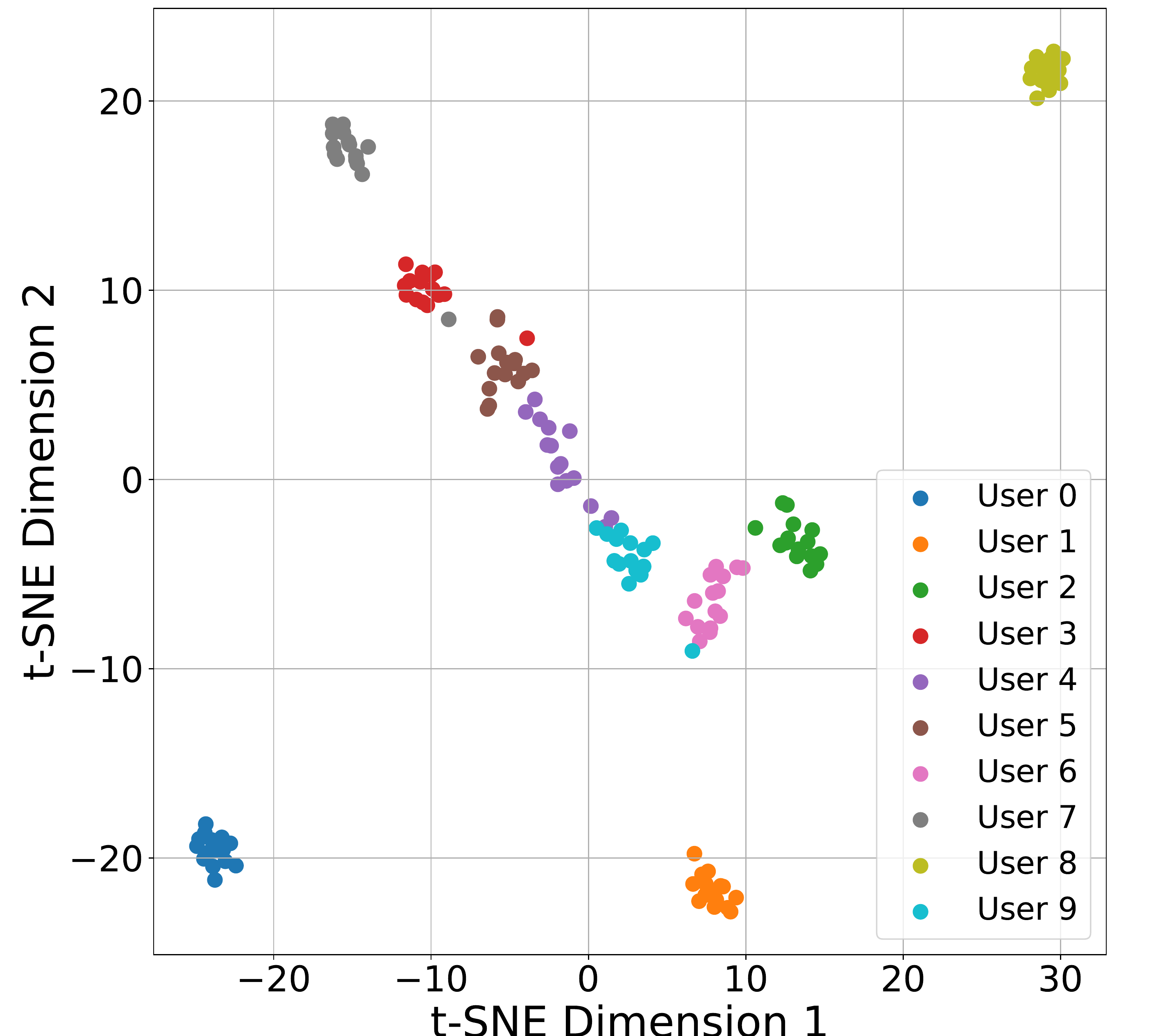}
    % \caption{}     
  \caption[]{2D graphical visualisation of the latent space through t-SNE considering 15 sessions of 10 subjects \cite{van2008visualizing}. Selected parameters\footnotemark: $\texttt{perplexity} = 14$, $\texttt{init} = \texttt{'pca'}$, $\texttt{n\_iter} = 1000$.}
    \label{fig:t-sne}
\end{figure}

\subsection{Analysis of the Feature Embeddings}
\label{subsec:Analysis_Feature_Embeddings}
The output feature embeddings extracted by TypeFormer lie in a 64-dimensional space and their pairwise relative positioning is measured throughout the Euclidean distance. In this scenario, mathematical methods like the popular t-SNE \cite{van2008visualizing} are useful to visualise data points in such high-dimensional spaces. Fig. \ref{fig:t-sne} depicts the output feature embedding space reduced to two dimensions through t-SNE. For better visualisation, we include examples of 10 random subjects of the database (15 acquisition sessions per subject). Apart from few outliers, most groups are clearly separated, while data points belonging to the same subjects are closer together. This is an indicator of small intra-class variability, and high inter-class variability.

%In the graph on the left-hand side, 15 output embeddings from 10 different users are depicted. In this case, apart from few outliers most of the embeddings of each of the users are clearly separated. In the plot on the right-hand side, 15 output embeddings of a single user are grouped considered the remaining distribution of the remaining 999 users. Also in this case, the embeddings belonging to one user are effectively grouped.

\footnotetext{\texttt{\href{https://scikit-learn.org/stable/modules/generated/sklearn.manifold.TSNE.html}{sklearn.manifold.TSNE -- scikit-learn 1.1.1}} documentation.\iffalse Accessed: 24-11-2022.\fi}

% Finally, the proposed Transformer outperforms the state-of-the-art approaches in the Aalto Database \cite{acien2021typenet}. It is important to highlight that both studies have the same experimental protocol.

% \subsection{Discussion}
% \label{subsec:Discussion}
% {\color{blue}

% The system performance improvement achieved with our proposed Transformer in relation with previous approaches is due, in our opinion, to the following reasons: \textit{(i)} our model applies the self-attention mechanism, being able to operate over long-distances in the input sequence; \textit{(ii)} our model attends to all the prior samples of the time sequence at the same time, without summarising the previous seen information; \textit{(iii)} the features are extracted from two different perspectives, from the time and the channel modules, providing more complex information; and \textit{(iv)} the Gaussian range encoding together with the multi-scale keystroke CNN allow to obtain a perspective of each sample in different environments, as different ranges are treated at the same time.
% }

\section{Conclusions and Future Work}
\label{sec:Conclusions_and_Future_Work}
In the current article, we have proposed a novel Transformer-based architecture, TypeFormer, for the task of free-text mobile keystroke authentication. TypeFormer features two branches (Temporal and Channel Modules) with Long Short-Term Memory (LSTM) layers, Gaussian Range Encoding (GRE), a multi-head Self-Attention mechanism, a Block-Recurrent Transformer layer, and it was trained with triplet loss. Its output consists in feature embedding vectors representing points in the output hyper-space. The distance between embedding vectors is measured through the Euclidean distance and it is less for instances of data belonging to the same subject than for ones of different subjects. The development of the model is based on the Aalto mobile keystroke database \cite{palin2019people}, the largest public databases of mobile keystroke dynamics. First, we have performed an analysis to validate the different modules that are present in the final presented Transformer architecture. Then, in order to compare TypeFormer with the highest-performing systems recently proposed in the literature, we have replicated the experimental protocol of two recent studies \cite{acien2021typenet, StragapedeMobile}, by varying the number of enrolment sessions ($E = 1, 2, 5, 7, 10$), input keystroke sequence lengths ($L = 30, 50, 70, 100$), and considering the same database repartition. In all cases, TypeFormer outperformed previous approaches, reaching as little as 3.25\% EER considering $E = 5$ and $L = 50$. This would be an absolute improvement of 5.95\% EER with respect to previous LSTM RNN-based model (the corresponding relative improvement is around 65\%) \cite{acien2021typenet}.  Moreover, we have assessed the ability of TypeFormer to model heterogeneous data and to extract robust features by considering other public databases for evaluation purposes only. The results obtained show a higher effectiveness of the proposed system in comparison to existing ones.
To advance the state of the art of free-text mobile keystroke biometrics, we make our proposed approach and experimental framework public\footnote{\texttt{\url{https://github.com/BiDAlab/TypeFormer}}}.
\par Concerning future work, the next directions of research will go towards exploring the effectiveness of Transformers in modelling other biometric traits \cite{TOLOSANA2022108609}, including data captured by mobile device sensors \cite{STRAGAPEDE2023109089, StragapedeIJCB}. To this end, we will consider the optimisation of the Transformer architecture to improve the performance with longer sequences. Additionally, more sophisticated training approaches will be investigated, in terms of the loss function, such as the implementation of hard triplet mining, in order to force the model to learn from harder comparisons \cite{Schroff_2015_CVPR}, and output feature embedding distance metrics. Finally, t would also be interesting to shed light on privacy aspects of mobile keystroke authentication, i.e., investigating the subject information contained in the feature embeddings, i.e., gender, age, etc., to assess whether keystroke data should be treated as privacy-sensitive biometric data. For this, the Aalto mobile keystroke database can be useful due to the the subject metadata available.

\section*{Acknowledgments}
This project has received funding from the European Union’s Horizon 2020 research and innovation programme under the Marie Skłodowska-Curie grant agreement No. 860315. Moreover, R. Tolosana and R. Vera-Rodriguez are also supported by INTER-ACTION (PID2021-126521OB-I00 MICINN/FEDER).

\bibliographystyle{IEEEtran}
\bibliography{0_Main}

\newpage

 

% \end{IEEEbiography}



% \vfill

\end{document}